\documentclass[journal]{IEEEtran}
\usepackage{cite}
\usepackage[numbers,sort&compress]{natbib}
\usepackage{graphicx}
\usepackage{dblfloatfix}  
\usepackage{epstopdf}
\usepackage{leftidx}
\usepackage{amsmath,amsfonts,amsthm}
\usepackage{bm} 
\usepackage{booktabs} 
\usepackage{caption}
\usepackage[colorlinks,
            linkcolor=black,
            anchorcolor=black,
            citecolor=black
            ]{hyperref}
\usepackage[nameinlink]{cleveref}
\Crefname{figure}{Figure.}{Figures.}

\begin{document}
\renewcommand{\thetable}{\arabic{table}}
\captionsetup[figure]{name={Figure.}}
\captionsetup[table]{name={Table.}}
%
\title{Linked Dynamic Graph CNN: Learning on Point Cloud via Linking Hierarchical Features}

\author{Kuangen Zhang$^{1,2}$, Ming Hao$^{3}$, Jing Wang$^{2}$, Clarence W. de Silva$^{2}$, and Chenglong Fu$^{1,*}$
\thanks{$^{1}$ K. Zhang, C. Fu are with the Department of Mechanical and Energy Engineering, Southern University of Science and Technology, Shenzhen 518055, China (Corresponding author: Chenglong Fu: fucl@sustc.edu.cn).}
\thanks{$^{2}$ K. Zhang, J Wang, C. W. de Silva are with the Department of Mechanical Engineering, The University of British Columbia, Vancouver V6T1Z4, Canada.}
\thanks{$^{3}$ M. Hao is with the Department of Mechanical Engineering, Tsinghua University, Beijing 100084, China.}
\thanks{Code and data: \url{https://github.com/KuangenZhang/ldgcnn}}
}

\markboth{}%
{Shell \MakeLowercase{\textit{et al.}}: Bare Demo of IEEEtran.cls for IEEE Journals}

\maketitle

\begin{abstract}
Learning on point cloud is eagerly in demand because the point cloud is a common type of geometric data and can aid robots to understand environments robustly. However, the point cloud is sparse, unstructured, and unordered, which cannot be recognized accurately by a traditional convolutional neural network (CNN) nor a recurrent neural network (RNN). Fortunately, a graph convolutional neural network (Graph CNN) can process sparse and unordered data. Hence, we propose a linked dynamic graph CNN (LDGCNN) to classify and segment point cloud directly in this paper. We remove the transformation network, link hierarchical features from dynamic graphs, freeze feature extractor, and retrain the classifier to increase the performance of LDGCNN. We explain our network using theoretical analysis and visualization. Through experiments, we show that the proposed LDGCNN achieves state-of-art performance on two standard datasets: ModelNet40 and ShapeNet.
\end{abstract}
\begin{IEEEkeywords}
Deep learning, Graph CNN, point cloud, classification, segmentation.
\end{IEEEkeywords}

\IEEEpeerreviewmaketitle

\section{Introduction}
\label{sec:Introduction}
\IEEEPARstart{T}hree dimensional (3D) perceptions aid robots to perceive and understand environments, thus increase the intelligence of robots. 3D geometric data can be provided by the depth camera, LiDAR scanners, and Radars. These sensors usually measure the distance between the target and sensor by projecting impulse signals to the target, such as laser and infrared light,  and measuring the reflected pulses or time of flight. Hence, 3D geometric data can provide 3D spatial information and are less affected by the intensity of illumination, which are more robust than RGB (red, green, and blue) images. For instance, 3D sensors can perceive environments at night. Besides, 3D geometric data from different viewpoints can be fused to provide complete environmental information. Consequently, 3D geometric data are crucial for robots to execute tasks in real environments. Additionally, 3D understanding is eagerly in demand in real applications, including self-driving \cite{qi_frustum_2017}, automatic indoor navigation \cite{zhu_target-driven_2017}, and robotics \cite{zhang_environmental_2019, zhang_sensor_2019}. 

A point cloud generally represents a set of 3D points in 3D space ($\mathbb{R}^3$). Each point in the point cloud has three coordinates: $x$, $y$, and $z$. The point cloud is the most common representation of 3D geometric data. For instance, the raw data of depth camera, LiDAR scanners, and Radars, are usually point clouds. Moreover, other 3D representations of geometric data, such as the mesh, voxel, and depth image, can be converted to point clouds easily. Therefore, it is significant to recognize point clouds.

\begin{figure}[htpb]
\centerline{\includegraphics[width=8cm]{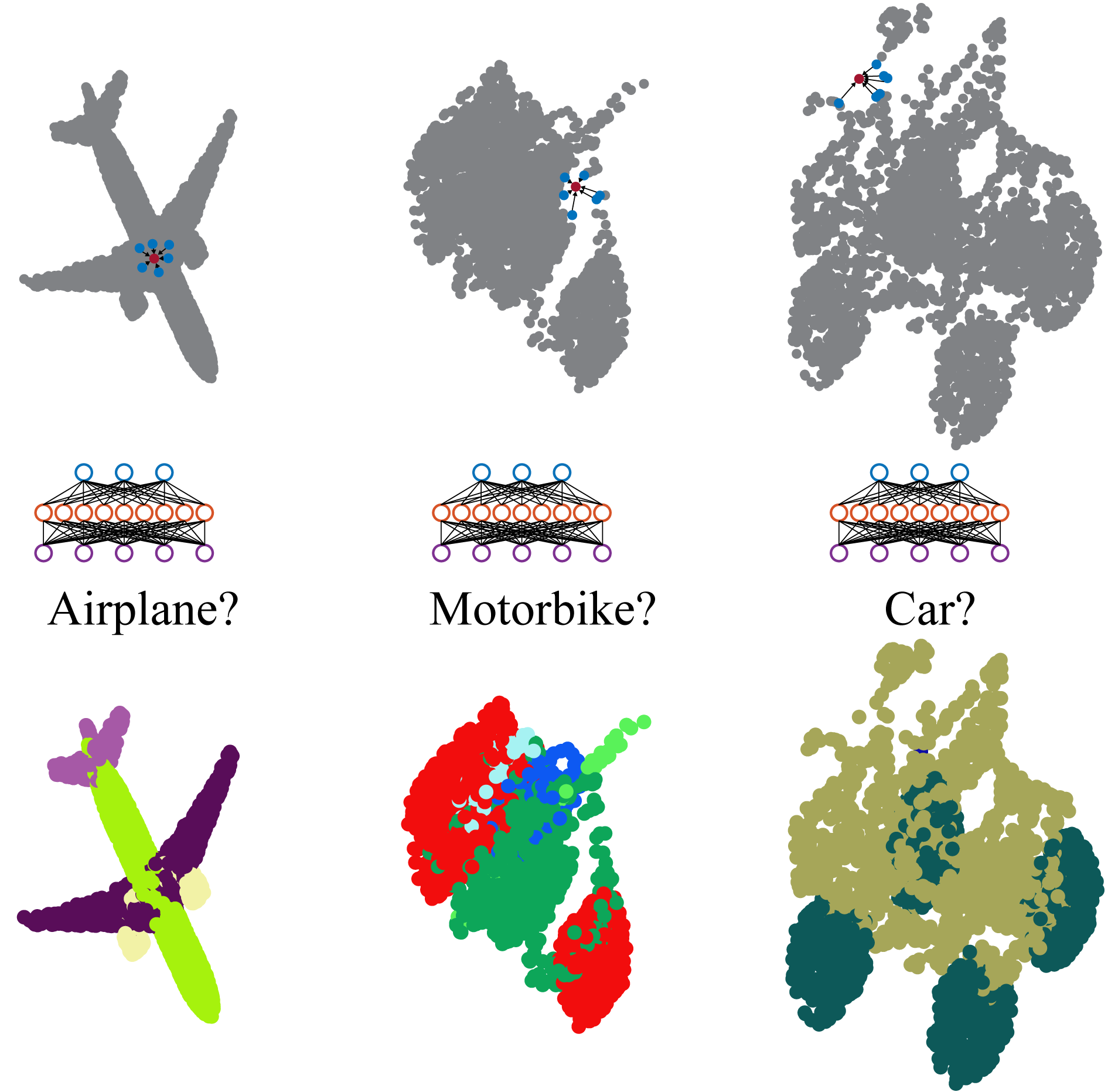}}
\caption{\textbf{Applications of LDGCNN}: point cloud classification and segmentation.}
\label{fig:1-Overview}
\end{figure}

For this reason, we consider designing methods to recognize point clouds in this paper. To be specific, we focus on two tasks: point cloud classification and segmentation. As shown in \autoref{fig:1-Overview}, the point cloud classification takes the whole point cloud as input and output the category of the input point cloud. The segmentation is to classify each point to a specific part of the point cloud. For these two tasks, there are two classic public datasets: ModelNet40 \cite{wu_3d_2015} and ShapeNet \cite{yi_scalable_2016}. A brief explanation of these two datasets is shown in \autoref{sec:Materials}.

Traditionally, researchers tend to design handcrafted features, like shape context \cite{belongie_shape_2002}, point feature histograms \cite{rusu_fast_2009}, local surface feature description \cite{guo_3d_2014}, and patch context \cite{hu_semantic_2018} to recognize point cloud. However, these handcrafted features are usually designed to solve specific problems and difficult to generalize to new tasks. Recently, deep learning methods achieve success on the classification and segmentation of objects \cite{krizhevsky_imagenet_2012, ronneberger_u-net:_2015}. Researchers are inclined to design data-driven methods to learn features and recognize point cloud automatically.

Deep learning methods have developed significantly in processing 2D images \cite{girshick_rich_2014}, but it is still challenging to classify and segment point cloud accurately. Because point cloud is sparse, unstructured, and unordered, typical convolutional neural network (CNN) and recurrent neural network (RNN), which require regular image format or ordered sequence, are not suitable to recognize point cloud directly. To recognize point cloud, previous researchers propose three kinds of deep learning methods: view based method, volumetric method, and geometric method. We discuss these methods in detail below.

\subsection{View based method}
Previous researchers attempt to project a point cloud to 2D images along different directions and apply standard 2D CNN to extract features. The features learned from different images are aggregated to a global feature through a view-pooling layer, then this global feature can be utilized to classify objects \cite{wu_3d_2015, yavartanoo_spnet:_2018, chen_veram:_2018}. Although the view based method can achieve high accuracy in the classification task, it is nontrivial to apply this method to segment point cloud, which classifies each point to a specific category.

\subsection{Volumetric method}
Another method to recognize the point cloud is to apply voxelization to convert the unstructured point cloud to a structured 3D grid. Then 3D CNN and volumetric CNN can be utilized to classify and segment the 3D grid \cite{maturana_voxnet:_2015, wang_voxsegnet:_2019}. Nevertheless, the point cloud is sparse, and it is wasteful to use a 3D grid to represent a point cloud. Additionally, considering the high memory and computational cost of the volumetric data, the resolution of the 3D grid is usually low, which can cause quantization artifacts. Consequently, it is problematic to utilize the volumetric method to process the large-scale point cloud.

\subsection{Geometric method}
Recent years, Qi et al. introduce a PointNet to classify and segment point cloud directly \cite{qi_pointnet:_2017}, which pioneers the geometric method to process unstructured data \cite{zhang_directional_2019}. There are several characteristics for point cloud: sparsity, permutation invariance, and transformation invariance. Considering the sparsity of point clouds, researchers of PointNet process points directly rather than projecting the point cloud to images or a volumetric grid. To solve the permutation invariant problem, they design a multi-layer perceptron (MLP) to extract features from each point independently. Because MLPs for different points share the parameters, the same type of features can be extracted from different points. Moreover, they use the max pooling layer to extract a global feature to aggregate the information from all points. Both the shared MLP and max-pooling layer are symmetric, and thus the permutation invariant problem is solved. Additionally, they design a transformation network to estimate the affine transformation matrix of the point cloud. Then they offset the point cloud using the estimated transformation matrix to solve the transformation invariant problem. 

PointNet is ingenious and can classify and segment point cloud directly, but it processes each point individually without extracting the local information between a point and its neighbors. Local features among neighboring points can be more robust than the coordinates of each point because the point cloud can rotate and shift. In consequence, researchers of PointNet improve their network to PointNet++ \cite{qi_pointnet++:_2017}. They apply PointNet recursively on the nested partitions to extract local features and combine learned features from multiple scales. After extracting the local feature, PointNet++ achieves the state-of-art results for point cloud classification and segmentation tasks on several common 3D datasets. However, PointNet++ still processes each point in the local point set individually and does not extract the relationships, such as distance and edge vector, between the point and its neighbors.

Recently, researchers start to design variants of PointNet for learning local features from the relationships of point pairs. The k-dimensional tree (KD-tree) is utilized to split point cloud into subdivisions. Researchers perform the multiplicative transformations on these subdivisions step by step and classify the point cloud. However, these partitions based on the KD-tree can vary if the scale of point cloud changes. Wang et al. design an edge convolutional operator to extract feature from a center point ($\bm{p_c}$) and the edge vector from its neighbor to itself ($\bm{p_n}-\bm{p_c}$). Moreover, they apply the k-nearest neighbors (K-NN) algorithm before each edge convolutional layer. Hence, not only do they search neighbors in the input Euclidean space $\mathbb{R}^3$, but also they cluster similar features in the feature space. Benefiting from extracting dynamic features, their dynamic graph CNN (DGCNN) achieves state-of-art results on several point cloud recognition tasks \cite{wang_dynamic_2018}. Lately, Xu et al. state that MLP does not work well on point cloud and they design a new convolutional layer, SpiderCNN, based on the Taylor polynomial to extract local geodesic information \cite{xu_spidercnn:_2018}. However, for the same input (1024 points) as DGCNN, the classification accuracy of SpiderCNN is slightly lower than that of DGCNN. Hence, this complex convolutional kernel seems not to outperform the MLP, which is concise and effective. To extract hierarchical features from the point cloud, Li et al. downsample the point cloud randomly and apply PointCNN to learn relationships among new neighbors in sparser point cloud \cite{li_pointcnn:_2018}. Moreover, they learn a transformation matrix from the local point set to permutate points into potentially canonical order. They decrease the forward time of network because of downsampling and achieves similar accuracy as DGCNN. However, the downsampling may influence the accuracy of the segmenting point cloud because each point should be classified into a category.

Up to now, the DGCNN achieves the highest classification accuracy on the modelNet40 dataset when there are only 1024 points in the input point cloud. Although some other methods, such as self-organizing network \cite{li_so-net:_2018} and Geometric CNN (Geo-CNN) \cite{lan_modeling_2018}, can increase the classification accuracy further, they require denser point cloud, which consists of 10,000 or 5,000 points and normal vectors. Because datasets of ModelNet40 and ShapeNet are generated from computer-aided design (CAD) models, it is easy to acquire denser point cloud with normal vectors. Nevertheless, it is a different case in real applications. 3D sensors, like LiDAR scanners, can only capture sparse point cloud. Moreover, there is no normal vector in the raw data of 3D sensors. Consequently, DGCNN seems to be more practical, but there are also some problems. Firstly, the DGCNN relies on the transformation network to offset the point cloud, but this transformation network doubles the size of the network. Besides, the deep features and their neighbors may be too similar to provide valuable edge vectors. Moreover, there are many trainable parameters in the DGCNN, and it is difficult to find the best parameters when we train the whole network.

In this paper, we optimize the network architecture of DGCNN to increase the performance and decrease the model size of the network. Because our network links the hierarchical features from different dynamic graphs, we call it linked dynamic graph CNN (LDGCNN). We apply K-NN and MLP with sharing parameters to extract the local feature from the central point and its neighbors. Then we add shortcuts between different layers to link the hierarchical features to calculate informative edge vectors. Moreover, there are two parts in our LDGCNN: convolutional layers (feature extractor) and fully connected layers (classifier). After training the LDGCNN, we freeze the feature extractor and retrain the classifier to improve the performance of the network. In the experiments, we evaluate our LDGCNN on two public datasets: ModelNet40 and ShapeNet for classification and segmentation. The experimental results show that our LDGCNN achieves state-of-art performance on these two datasets.

The key contributions of this paper include:
\begin{itemize}
    \item We link hierarchical features from different dynamic graphs to calculate informative edge vectors and avoid vanishing gradient problem.
    \item We remove the transformation network from the DGCNN and demonstrate that we can use MLP to extract transformation invariant features.
    \item We increase the performance of LDGCNN by freezing the feature extractor and retraining the classifier. 
    \item We evaluate our LDGCNN and show that it achieves state-of-art performance on two classic 3D datasets.
\end{itemize}

In the following parts of this paper, we introduce the dataset of ModelNet40 and ShapeNet in \autoref{sec:Materials}. Then we describe the research problems, our theoretical methods, and network architecture in \autoref{sec:Methods}. The experimental results and corresponding discussions are shown in \autoref{sec:Results}. Finally, we conclude this paper in \autoref{sec:Conclusions}.

\section{Materials}
\label{sec:Materials}
The dataset of classification is ModelNet40 \cite{wu_3d_2015}, which includes 12,311 CAD modes belonging to 40 categories. This dataset is split into training set (9843 models) and validation set (2468 models) \cite{qi_pointnet:_2017}. However, there is no testing set for ModelNet40. We discuss this problem in \autoref{subsubsec: classification}. In addition, each CAD model is sampled by 1024 points, which are normalized to a unit sphere. 

The segmentation task is based on ShapeNet part dataset \cite{yi_scalable_2016}, which consists of 
16,881 CAD models from 16 categories. Previous researchers sample each CAD model to 2048 points, and each point is annotated with one of 50 parts. The training, validation, and testing datasets in our paper are the same as in \cite{wang_dynamic_2018}.

\section{Methods}
\label{sec:Methods}
Our LDGCNN is inspired by PointNet \cite{qi_pointnet:_2017} and DGCNN \cite{wang_dynamic_2018}. We construct a directed graph for the point cloud. Then we extract features from the constructed graph and utilize features to classify and segment point cloud. In the following, we discuss the research problems and corresponding solutions.

\subsection{Problem statement}
The input of our method is a point cloud, which is a set of 3D points:
\begin{equation}
    \{\bm{p_i} = (x_i, y_i, z_i)| i = 1, 2,...,n\}
\end{equation}
where $\bm{p_i}$ is one point of point cloud and consists of three coordinates $(x_i, y_i, z_i)$. The index of this point and the number of points in the point cloud are $i$ and $n$, respectively.

We focus on two tasks in this paper: point cloud classification and segmentation. For point cloud classification, we need to classify the category of the whole point cloud. Hence, we should find a classification function $f_c$ to convert the input point cloud to the probability distribution on each category $\bm{Pr}$:
\begin{equation}
    \bm{Pr} = f_c(\{\bm{p_1}, \bm{p_2},..., \bm{p_n}\})
\end{equation}

As for the point cloud segmentation, each point $\bm{p_i}$ can be classified as a specific category. Therefore, we need to find a segmentation function $f_s$ to calculate the probability distribution on each category $\bm{Pr_i}$ for each point $\bm{p_i}$:
\begin{equation}
    \{\bm{Pr_i}|i = 1, 2,..., n\} = \{f_s(\bm{p_i})|i = 1, 2,..., n\}
\end{equation}

There are several design constraints for the classification function $f_c$ and segmentation function $f_s$;
\subsubsection{Permutation invariance}
\label{subsubsec:permutation_invariance}
The point cloud is a point set rather than the sequence signal. Thus the order of points may vary but does not influence the category of the point cloud. The classification function should not be influenced by order of input points:
\begin{equation}
\begin{split}
    (j_1, j_2,..., j_n) &\neq (i_1, i_2,..., i_n)\\
    f_c(\{\bm{p_{j_1}}, \bm{p_{j_2}},..., \bm{p_{j_n}}\}) &= f_c(\{\bm{p_{i_1}}, \bm{p_{i_2}},..., \bm{p_{i_n}}\})
\end{split}
\end{equation}
where $(j_1, j_2,..., j_n)$ and $(i_1, i_2,..., i_n)$ represent two different sequences.

\subsubsection{Transformation invariance}
\label{subsubsec:transformation_invariance}
In the real application, the relative position and direction between the sensor and objects might change, thus causing the translation and rotation of generated point cloud. However, the results of classification and segmentation should not be changed by the above affine transformations:
\begin{equation}
\begin{split}
    f_c(\bm{R}\{\bm{p_{i_1}}, \bm{p_{i_2}},..., \bm{p_{i_n}}\} + \bm{b}) &= f_c(\{\bm{p_{i_1}}, \bm{p_{i_2}},..., \bm{p_{i_n}}\})\\
    f_s(\bm{R}\bm{p_i} + \bm{b}) &=  f_s(\bm{p_i})
\end{split}
\end{equation}
where $\bm{R}$ and $\bm{b}$ are rotation matrix and translation vector.

\subsubsection{Extracting local features}
\label{subsubsec:local_features}
Local features are relationships between the point and its neighbors, which are critical to the success of point cloud recognition. Consequently, we need to learn the local features rather than process each point individually:
\begin{equation}
\begin{split}
    \bm{P} &= \{\bm{p_i} = (x_i, y_i, z_i)| i = 1, 2,...,n\} \subset \mathbb{R}^3\\
    \bm{L_i} &= \{\bm{p_{i_k}}| k = 1, 2,..., K\} \subseteq \bm{P}\\
    \bm{Pr} &= f_c(\{\bm{L_1}, \bm{L_2},..., \bm{L_n}\})\\
    \bm{Pr_i} &= f_s(\bm{L_i})
\end{split}
\end{equation}
where  $\bm{L_i}$ and $\bm{p_{i_k}}$ are the local point set and neighbor of a point $\bm{p_i}$.

\subsection{Graph generation}
Graph neural network is an applicable method to process point cloud because it propagates on each node individually ignoring the input order of nodes and extracts edges to learn the information of dependency between two nodes \cite{zhou_graph_2018}. To apply graph neural network on the point cloud, we need to convert it to a directed graph first. Graph $G$ is composed of vertices $V$ and edges $E$:
\begin{equation}
\begin{split}
    G = (V, E), V \subset \mathbb{R}^C, E \subseteq V \times V
\end{split}
\end{equation}
where $C$ is the dimension of each vertice.

\begin{figure}[htbp]
\centerline{\includegraphics[width=8cm]{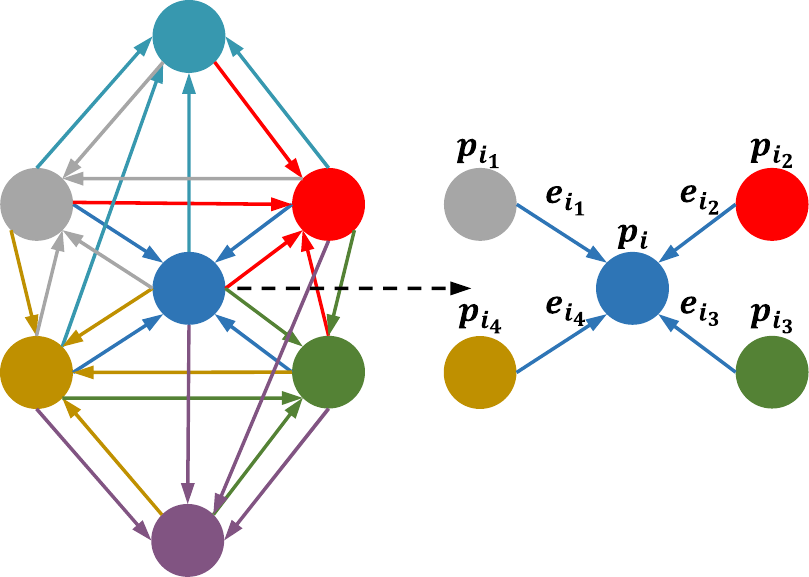}}
\caption{\textbf{Graph of point cloud}. The $\bm{p_i}$ and $\{\bm{p_{i_1}},..., \bm{p_{i_4}}\}$ are a central point and its neighbors. The directed edges from the neighbors to the central point are denoted by $\{\bm{e_{i_1}},..., \bm{e_{i_4}}\}$.}
\label{fig:2-graph}
\end{figure}

We do not construct fully connected edges for the point cloud because it consumes large memory. A simple method is to utilize K-NN to construct a locally directed graph, which is shown in \autoref{fig:2-graph}. In this local directed graph, each point $\bm{p_i}$ is a central node, and the edges $\bm{e_i}$ between the central node and its k-nearest neighbors are calculated:
\begin{equation}
\begin{split}
    G &= (V, E)\\
    V &= \{\bm{p_i}| i = 1, 2,..., n\}\\
    E &= \{\bm{e_i} = (\bm{e_{i_1}}, \bm{e_{i_2}},..., \bm{e_{i_K}})| i = 1, 2,..., n\}\\
    \bm{e_{i_j}} &= \bm{p_{i_j}} - \bm{p_i}
\end{split}
\end{equation}
where $\bm{p_{i_j}}$ and $\bm{e_{i_j}}$ are one neighbor of central point $\bm{p_i}$ and the directed edge from the neighbor $\bm{p_{i_j}}$ to the cnetral point $\bm{p_i}$, respectively. The number of neighbors is denoted by $K$.

Like DGCNN \cite{wang_dynamic_2018}, we apply K-NN before each convolutional layer, then we can construct the local graph in both Euclidean space and feature space. Hence, the point $\bm{p_i}$ also represents the central point in the feature space and the point $\bm{p_{i_j}}$ is a neighbor of central point $\bm{p_i}$ in the feature space.

\subsection{Graph feature extraction}
After constructing the local graph, we extract local graph features based on the edge convolutional layer \cite{wang_dynamic_2018}. The feature extraction function $f_e$ is the same for all points, hence we exemplify this operation using one central point $\bm{p_i}$  and its $K$ neighbors. The input is the local graph of the central point $\bm{p_i}$ and the output is the local feature $\bm{l_i}$:

\begin{equation}\label{eq:local_feature}
\begin{split}
    \bm{l_i} &= f_e(G(\bm{p_i}, \bm{e_i}))\\
             &= \max\{\bm{h}(\bm{p_i}, \bm{e_{i_1}}), \bm{h}(\bm{p_i}, \bm{e_{i_2}}),..., \bm{h}(\bm{p_i}, \bm{e_{i_K}})\}
\end{split}
\end{equation}
where $\bm{h}(\bm{p_i}, \bm{e_{i_j}})$ is the hidden feature vector for the central point $\bm{p_i}$ and one edge vector $\bm{e_{i_j}}$.

In (\ref{eq:local_feature}), we use the max-pooling operation because it is not influenced by the order of neighbors and can extract the most predominant feature among all edges. Moreover, we utilize a MLP to extract the hidden feature vector $\bm{h}(\bm{p_i}, \bm{e_{i_j}})$:
\begin{equation}
\begin{split}
    h_{c'}(\bm{p_i}, \bm{e_{i_j}}) &= \sum_{c=1}^C w_{c'c}p_{ic} + w_{c'(c+C)}(p_{i_jc} - p_{ic}) + b_{c'}\\
    \bm{h(\bm{p_i}, \bm{e_{i_j}})} &= (h_{1}(\bm{p_i}, \bm{e_{i_j}}), h_{2}(\bm{p_i}, \bm{e_{i_j}}),..., h_{C'}(\bm{p_i}, \bm{e_{i_j}}))
\end{split}
\end{equation}
where $p_{ic}$ and $p_{i_jc}$ are values in the channel $c$ for the central point $\bm{p_i}$ and its neighbor $\bm{p_{i_j}}$. The numbers of channels for input point $\bm{p_i}$ and output hidden feature vector $\bm{h(\bm{p_i}, \bm{e_{i_j}})}$ are $C$ and $C'$, respectively. The value in the channel $c'$ for the hidden feature vector is denoted by $h_{c'}(\bm{p_i}, \bm{e_{i_j}})$. The trainable parameters for the MLP are $w_{c'c}$, $w_{c'(c+C)}$, and $b_{c'}$.

\subsection{Transformation invariant function}
\label{subsubsec:transformation}
Both PointNet \cite{qi_pointnet:_2017} and DGCNN \cite{wang_dynamic_2018} rely on a transformation network to estimate the rotation matrix of the point cloud and offset the point cloud. However, the transformation network doubles the size of the network. Moreover, through experiments, we find that the network still has satisfactory performance after removing the transformation network. Therefore, we discuss the corresponding reasons here.

The output of transformation network is a $3\times3$ matrix $\bm{R}$, which can offset the point cloud $\bm{P}$:
\begin{equation}\label{eq:rotaion_offset}
\begin{split}
    \bm{P}_\text{offset} = \begin{bmatrix}
    x_{1} & y_{1} & z_{1} \\
    x_{2} & y_{2} & z_{2} \\
    \vdots & \vdots & \vdots \\
    x_{n} & y_{n} & z_{n} \\
    \end{bmatrix}
    \begin{bmatrix}
    r_{11} & r_{12} & r_{13} \\
    r_{21} & r_{22} & r_{23} \\
    r_{31} & r_{32} & r_{33} \\
    \end{bmatrix}
\end{split}
\end{equation}
where $r_{ij}$ is the value of rotation matrix $\bm{R}$ on the $i$th row and $j$th column. Additionally, $x_i$, $y_i$, and $z_i$ are coordinates of one point $\bm{p_i}$ in the point cloud $\bm{P}$.

If we convert the MLP to matrix form, we can find the hidden feature vector is:
\begin{equation}\label{eq:MLP}
\begin{split}
    \bm{h}(\bm{P}) &= \begin{bmatrix}
    x_{1} & y_{1} & z_{1} \\
    x_{2} & y_{2} & z_{2} \\
    \vdots & \vdots & \vdots \\
    x_{n} & y_{n} & z_{n} \\
    \end{bmatrix}
    \begin{bmatrix}
    w_{11} & w_{12} & \cdots & w_{1C'} \\
    w_{21} & w_{22} & \cdots & w_{2C'} \\
    w_{31} & w_{32} & \cdots & w_{3C'}
    \end{bmatrix}\\
    &+ \begin{bmatrix}
    b_{1} & b_{2} & \cdots & b_{C'}
    \end{bmatrix} 
\end{split}
\end{equation}
where $w_{ic'}$ and $b_{c'}$ are trainable parameters for the MLP in the channel $c'$. The number of output channel is $C'$.

Compared (\ref{eq:MLP}) with (\ref{eq:rotaion_offset}), we can discover that they are similar. The difference is that the transformation network can estimate a specific matrix for each point cloud, whereas the MLP is static for all point cloud. The parameters of the MLP in different channels can rotate the point cloud with different angles and shift the point cloud with different translations. Considering that we have at least 64 output channels, the MLP can observe the point cloud along at least 64 directions, which can make the network to be rotation invariant approximately.

Furthermore, we augment the training data to increase the generalization capability of the network. Like the PointNet++ \cite{qi_pointnet++:_2017}, we also rotate, shift, and scale the input point cloud randomly and add random noise on each point. After augmenting the training data, the network can learn rotation and translation invariant features. 

\subsection{LDGCNN architecture}
\begin{figure*}[htpb]
\centerline{\includegraphics[width=16cm]{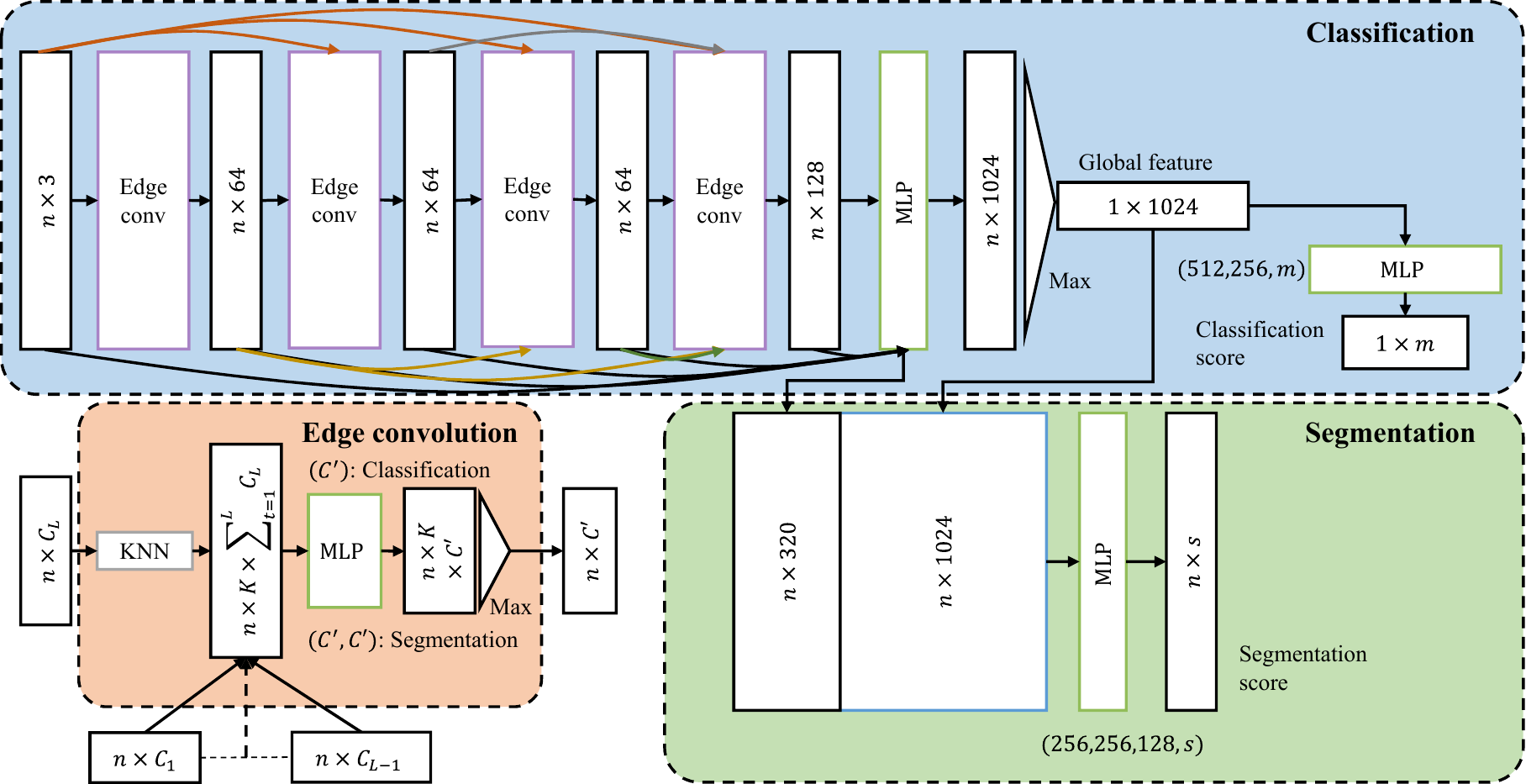}}
\caption{\textbf{LDGCNN architecture}. MLP represents the multi-layer perceptron, which shares parameters for all points and is a symmetric function. The layer sizes of MLP are the same as the sizes of output or indicated by the numbers in the bracket. In the edge convolutional layer, the MLP $(C', C')$ for segmentation has two layers with $C'$ channels.
The input of the classification network is a point cloud consisting of $n$ 3D points, and the output values are the classification scores for $m$ classes. The edge convolutional layer is applied to extract local features from the aggregated features and their neighbors, which consist of features from input layer (layer index = 1) to current layer (layer index = $L$). The numbers of channels in different layers are denoted by $C_1,..., C_L$. Then shortcuts and a max-pooling layer are used to aggregate features to a global feature, which is utilized to classify the point cloud by fully connected layers (MLPs). The segmentation network is an extension of classification network. We stack the global feature with the extracted local features and utilize MLPs to calculate segmentation scores for each point. We use the ReLU and Batchnorm in each layer. In the fully connected layers, Dropout layer is applied after each MLP layer.}
\label{fig:3-LDGCNN_architecture}
\end{figure*}

The architecture of our LDGCNN is shown in \autoref{fig:3-LDGCNN_architecture} and the detailed explanations are stated in the caption of \autoref{fig:3-LDGCNN_architecture}. Our network is similar to the DGCNN \cite{wang_dynamic_2018}, and there are several differences between our LDGCNN and DGCNN: 
\begin{itemize}
    \item We link hierarchical features from different layers.
    \item We remove the transformation network.
\end{itemize}

The reason for removing the transformation network is discussed in \autoref{subsubsec:transformation}. Here we explain why we link hierarchical features from different layers. The first reason is that linking hierarchical features can avoid vanishing gradient problem for the deep neural network, which is demonstrated in \cite{he_deep_2016}. Besides, neighborhood indices based on current features are different from that based on previous features. Hence, we can learn new features by using current indices to extract edges from previous features. Moreover,  neighbors of current features are similar to each other, which may cause the edge vector to approach zero. If we use current neighborhood indices to calculate the edges from previous features, we may get informative edge vectors.

\subsection{Freezing feature extractor and retraining classifier}
We increase the performance of our network further by freezing feature extractor (network before the global feature) and retraining the classifier (MLP after the global feature). Theoretically, training the whole network can find the global minimum. However, there are too many parameters in the network, and it is almost impossible to find the global minimum through training. Conversely, the network may fall into an inferior local minimum because of the large size of parameters. After freezing the feature extractor, the back propagation only influences the parameters of the classifier, which consists of three fully connected layers. Consequently, I save the global features after training the whole network, then I utilize these global features to train the classifier separately, which help my network to achieve better performance.

\section{Results}
\label{sec:Results}
\subsection{Implementation details}
Our training strategy is the same as \cite{wang_dynamic_2018}. The Adam optimizer with 0.001 learning rate is used to optimize the whole network. The number of input points, batch size, and momentum are 1024, 32 and 0.9, respectively. We set the number of neighbors $K$ to 20. Moreover, the dropout rate is set as 0.5 for all dropout layers. After training the whole network, I freeze the feature extractor and retrain the classifier. In the retraining process, I only change the optimizer to Momentum. For the segmentation network, we change the number of neighbors $K$ to 30 because the number of input points changes to 2048.  Other settings are the same as that for classification.

Furthermore, we train our classification network on one NVIDIA TITAN V GPU. Then we use two NVIDIA TITAN V GPUs to train our segmentation network.

\subsection{Point cloud classification}
\label{subsubsec: classification}
We evaluate the performance of our classification network on the ModelNet40 dataset \cite{wu_3d_2015}, which is introduced in \autoref{sec:Materials}. As shown in \autoref{tab:classification}, our LDGCNN achieves the highest accuracy on this dataset. For the same input (1024 points), the overall classification accuracy of our network (92.9\%) is 0.7\% higher than that of DGCNN and PointCNN. 

Previous researchers present that they split the ModelNet40 to the training set and testing set \cite{qi_pointnet:_2017, wang_dynamic_2018}. However, they write their best testing accuracy in their papers. The testing accuracy of the last epoch for their training can be about 1\%-1.5\% lower than their best testing accuracy. Consequently, they regard the testing set as the validation set and use the validation set to optimize their hyper-parameters and determine the time of stop training. Because we need to compare our networks with previous networks, we also use the validation set of ModelNet40 to determine the time to stop training. Since we use the same strategy of splitting dataset and training as previous researchers, the higher classification accuracy of our network still validates that our network achieves the state-of-art performance.

\begin{table}[htbp]
\centering
\caption {\label{tab:classification} Classification results on ModelNet40. MA represents mean per-class accuracy, and the per-class accuracy is the ratio of the number of correct classifications to that of the objects in a class.
OA denotes the overall accuracy, which is the ratio of the number of overall correct classifications to that of overall objects.}
\renewcommand{\arraystretch}{1} 
\begin{center}
\begin{tabular}{l l l l}
\toprule
Method & Input & MA (\%) & OA (\%) \\ 
\midrule
PointNet \cite{qi_pointnet:_2017} & 1024 points & 86.0 & 89.2 \\
PointNet++ \cite{qi_pointnet++:_2017} & 5000 points+normal & - & 90.7 \\
KD-Net \cite{klokov_escape_2017} & 1024 points & - & 91.8 \\
DGCNN \cite{wang_dynamic_2018} & 1024 points & 90.2 & 92.2 \\
SpiderCNN \cite{xu_spidercnn:_2018} & 1024 points+normal & - & 92.4 \\
PointCNN \cite{li_pointcnn:_2018} & 1024 points & 88.1 & 92.2 \\
\midrule
\textbf{Ours} & 1024 points & \textbf{90.3} & \textbf{92.9} \\
\bottomrule
\end{tabular}
\end{center}
\end{table}

\subsection{Point cloud segmentation}
\begin{figure}[htpb]
\centerline{\includegraphics[width=8cm]{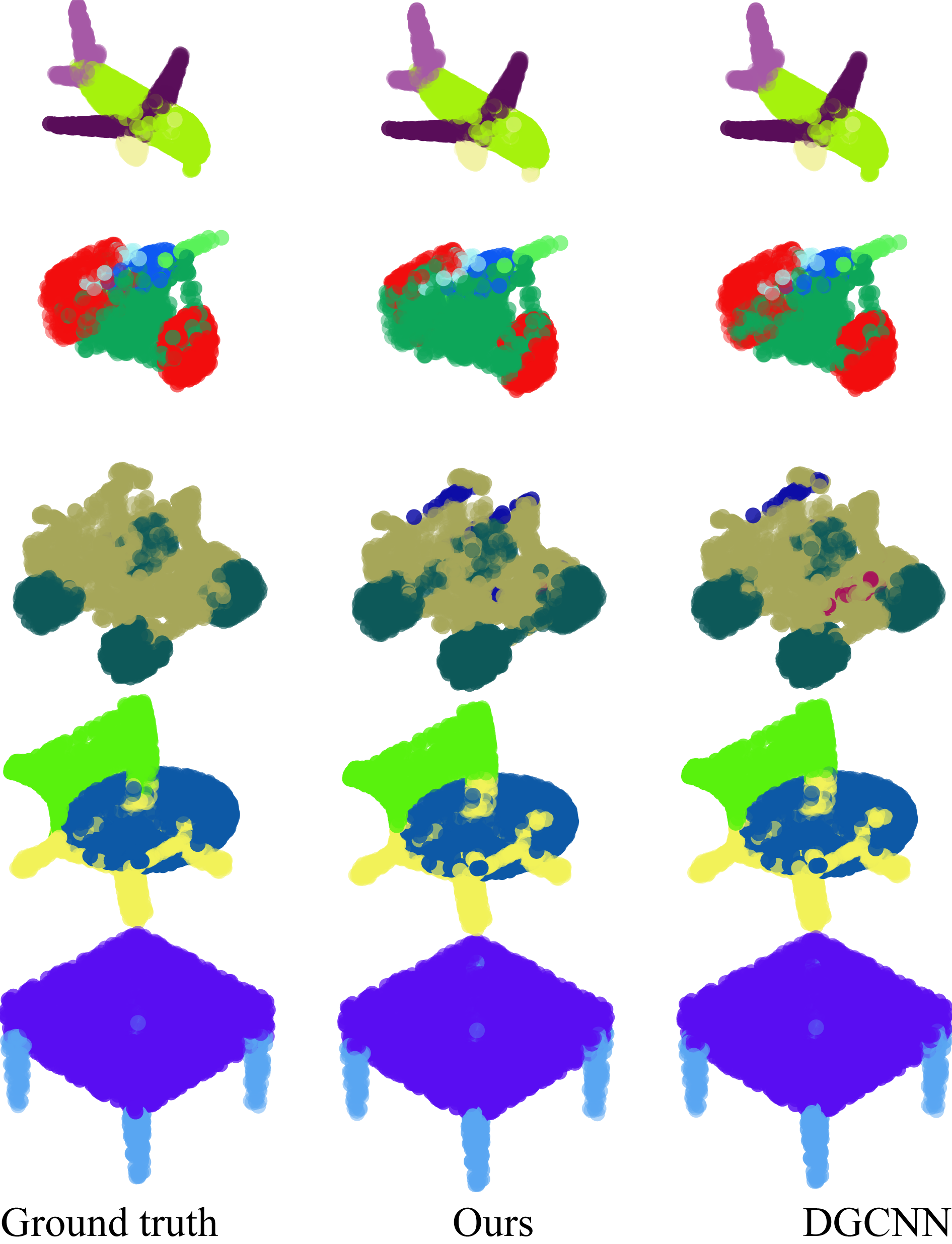}}
\caption{\textbf{Qualitative results for point cloud segmentation.} Comparison of segmentation results among ground truth (left), LDGCNN (middle), and DGCNN (right).}
\label{fig:4-segmentation}
\end{figure}

The Intersection-over-Union (IoU) is utilized to evaluate the performance of our network for point cloud segmentation on ShapeNet \cite{yi_scalable_2016}, which is described in \autoref{sec:Materials}. The segmentation network predicts the label of each point, and we compare the predictions with ground truth. Then the intersection and union of predicted points and ground truth points are calculated. The IoU is the ratio of the number of points in the intersection to that in the union. As shown in \autoref{tab:segmentation}, our segmentation network also achieves the state-of-art performance. Moreover, we compare our segmented point cloud with that of ground truth and DGCNN in \autoref{fig:4-segmentation}.

\begin{table*}[!t]
\centering
\caption {\label{tab:segmentation} Part segmentation results on ShapeNet. The values in the table are mean IoU (\%) on points}
\renewcommand{\arraystretch}{1} 
\begin{center}
\resizebox{2.1\columnwidth}{!}{
\begin{small}
\begin{sc}
\begin{tabular}{l|c|cccccccccccccccc}
\toprule
                & \textbf{mean} & areo & bag & cap & car & chair & ear & guitar & knife & lamp & laptop & motor & mug & pistol & rocket & skate & table\\
                & & &  &  & &  & phone&   & & & & & &  & & board & \\
\midrule
 \# shapes   &       &2690 &76  &55 & 898 & 3758 & 69 & 787 & 392 & 1547 & 451 & 202 & 184 & 283 & 66 & 152 & 5271\\
\midrule
PointNet \cite{qi_pointnet:_2017}  & 83.7 & 83.4 & 78.7 & 82.5 & 74.9 & 89.6 & 73.0 & \textbf{91.5} & 85.9 & 80.8 & 95.3 & 65.2 & 93.0 & 81.2 & 57.9 & 72.8 & 80.6\\
PointNet++ \cite{qi_pointnet++:_2017}  & \textbf{85.1} & 82.4 & 79.0 & \textbf{87.7} & 77.3 & 90.8 & 71.8 & 91.0 & 85.9 & \textbf{83.7} & 95.3 & \textbf{71.6} & 94.1 & 81.3 & 58.7 & \textbf{76.4} & \textbf{82.6}\\
KD-Net \cite{klokov_escape_2017}& 82.3 &  80.1 & 74.6 & 74.3 & 70.3 & 88.6 & 73.5 & 90.2 & 87.2 & 81.0 & 94.9 & 57.4 & 86.7 & 78.1 & 51.8 & 69.9 & 80.3\\
DGCNN \cite{wang_dynamic_2018}                     & \textbf{85.1} & \textbf{84.2} & \textbf{83.7} & 84.4 & 77.1 & \textbf{90.9} & \textbf{78.5} & \textbf{91.5} & 87.3 & 82.9 & \textbf{96.0} & 67.8 & 93.3 & \textbf{82.6} & \textbf{59.7} & 75.5 & 82.0\\
\midrule
\textbf{Ours} & \textbf{85.1} & 84.0 & 83.0 & 84.9 & \textbf{78.4} & 90.6 & 74.4 & 91.0 & \textbf{88.1} & 83.4 & 95.8 & 67.4 & \textbf{94.9} & 82.3 & 59.2 & 76.0 & 81.9\\

\bottomrule
\end{tabular}
\end{sc}
\end{small}
}
\end{center}
\end{table*}

\subsection{Time and space complexity analysis}
We evaluate our model complexity by comparing the number of parameters and forward time of our classification network with that of other classification networks. The forward time is estimated based on the computing capacity of one NVIDIA TITAN V GPU. As shown in \autoref{tab:time}, the model size of our network is smaller than that of PointNet, PointNet++, and DGCNN. The forward time of our network is shorter than that of DGCNN, but is longer than that of PointNet++ and PointNet, because we apply K-NN to search neighbors in each edge convolutional layer. Compared to DGCNN, our network is more concise and achieves higher classification accuracy.

\begin{table}[htbp]
\centering
\caption {\label{tab:time} Comparison of the number of parameters, forward time, and overall accuracy (OA) among different classification networks.}
\renewcommand{\arraystretch}{1} 
\begin{center}
\begin{tabular}{l l l l}
\toprule
Method & \#Parameters (M) & Forward time (ms) & OA (\%) \\ 
\midrule
PointNet \cite{qi_pointnet:_2017} & 3.48 & \textbf{0.8} & 89.2\\
PointNet++ \cite{qi_pointnet++:_2017} & 1.48 & 1.4 & 90.7 \\
DGCNN \cite{wang_dynamic_2018} & 1.84 & 3.1 & 92.2 \\
\midrule
\textbf{Ours} & \textbf{1.08} & 2.8 & \textbf{92.9} \\
\bottomrule
\end{tabular}
\end{center}
\end{table}

\subsection{Visualization and ablation experiments}
\begin{figure}[htpb]
\centerline{\includegraphics[width=8cm]{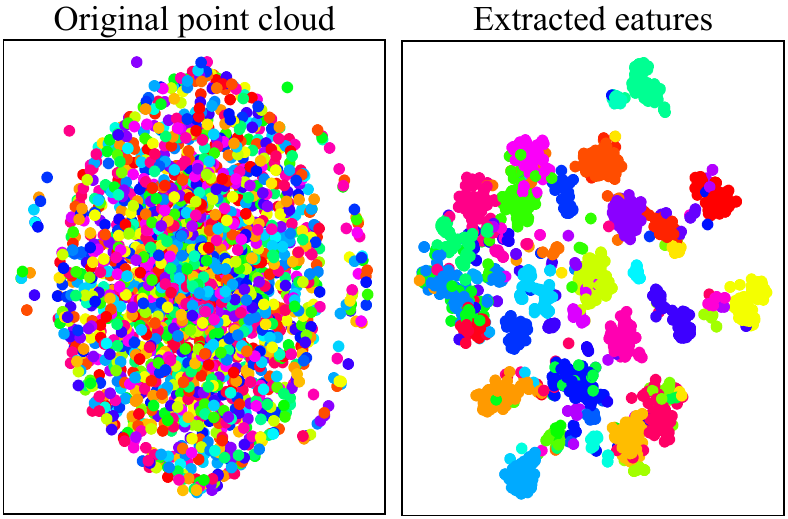}}
\caption{\textbf{T-SNE visualization of original point cloud and extracted features.} Points with the same color belong to the same category.}
\label{fig:5-t-SNE}
\end{figure}

T-distributed stochastic neighbor embedding (T-SNE) is utilized to show the performance of our feature extractor \cite{maaten_visualizing_2008}. The T-SNE reduces the dimension of high-dimensional features to visualize the separability of the features. As shown in \autoref{fig:5-t-SNE}, the extracted features are much more discriminative than original point cloud. 

Furthermore, we analyze the effects of our two components: removing the transformation network and freezing feature extractor and retraining the classifier. As shown in \autoref{tab:components}, we increase the overall classification accuracy from 91.8\% to 92.9\% by freezing feature extractor and retraining the classifier. Moreover, the transformation network cannot increase the performance of our network. This result validates our assumption that the MLP can substitute the transformation network for approximating rotation invariance, which is proposed in \autoref{subsubsec:transformation}.

\begin{table}[htbp]
\centering
\caption {\label{tab:components} Effects of combining different components. The definitions of MA and OA are the same as in \autoref{tab:classification}.}
\renewcommand{\arraystretch}{1} 
\begin{center}
\begin{tabular}{l | c c c c}
\toprule
Transformation network &  & x &   & x\\ 
Freezing \& retraining &  &   & x & x\\
OA(\%) & 91.8 & 91.8 & \textbf{92.9} & 92.4 \\
MA(\%) & 88.8 & 89.0 & \textbf{90.3} & 89.7 \\
\bottomrule
\end{tabular}
\end{center}
\end{table}
\section{Conclusions}
\label{sec:Conclusions}
In this paper, we proposed a LDGCNN to classify and segment point cloud directly. Compared to DGCNN, we decreased the model size of network by removing the transformation network and optimized the network architecture through linking hierarchical features from different dynamic graphs. After training the whole network, we froze the feature extractor and retrained the classifier, which increased the classification accuracy on ModelNet40 from 91.8\% to 92.9\%. Our network achieved state-of-art performance on two standard datasets: ModelNet40 and ShapeNet. In addition, we provided theoretical analysis of our network and visualized classification and segmentation results.
In the future, we will evaluate our LDGCNN on more semantic segmentation datasets and apply it on the real-time environmental understanding application.

\bibliographystyle{IEEEtran}
\bibliography{ldgcnn}

\end{document}